\documentclass[runningheads]{llncs}
\usepackage[utf8]{inputenc}
\usepackage{amsmath}
\usepackage{amsfonts}
\usepackage{algpseudocode}
\usepackage{graphicx}
\usepackage{url}
\usepackage[ruled]{algorithm2e}

\title{Best uses of ChatGPT and Generative AI for computer science research}
\author{Eduardo C. Garrido-Merchán}
\date{November 2023}

\institute{Universidad Pontificia Comillas, Madrid, Spain
\email{ecgarrido@icade.comillas.edu}}

\begin{document}

\maketitle

\begin{abstract}
Generative Artificial Intelligence (AI), particularly tools like OpenAI's popular ChatGPT, is reshaping the landscape of computer science research. Used wisely, these tools can boost the productivity of a computer research scientist. This paper provides an exploration of the diverse applications of ChatGPT and other generative AI technologies in computer science academic research, making recommendations about the use of Generative AI to make more productive the role of the computer research scientist, with the focus of writing new research papers. We highlight innovative uses such as brainstorming research ideas, aiding in the drafting and styling of academic papers and assisting in the synthesis of state-of-the-art section. Further, we delve into using these technologies in understanding interdisciplinary approaches, making complex texts simpler, and recommending suitable academic journals for publication. Significant focus is placed on generative AI's contributions to synthetic data creation, research methodology, and mentorship, as well as in task organization and article quality assessment. The paper also addresses the utility of AI in article review, adapting texts to length constraints, constructing counterarguments, and survey development. Moreover, we explore the capabilities of these tools in disseminating ideas, generating images and audio, text transcription, and engaging with editors. We also describe some non-recommended uses of generative AI for computer science research, mainly because of the limitations of this technology.
\end{abstract}

\section{Introduction}
The advent of Generative Artificial Intelligence (AI)  \cite{van2013generative}, epitomized by tools such as ChatGPT \cite{deng2022benefits} but also include a wide array of models  \cite{gozalo2023chatgpt,gozalo2023survey}, represents a paradigm shift in the domain of computer science research. These tools, leveraging advanced machine learning algorithms \cite{goodfellow2016deep}, have the potential to automate a wide array of tasks, not only creative writing \cite{garrido2023simulating} but, fundamentally, aiding research writing and transforming the workflow of computer science researchers. This paper sets out to provide an exhaustive exploration of the various applications of ChatGPT and similar generative AI technologies \cite{gozalo2023survey}, with a specific focus on enhancing the productivity of computer science researchers, especially in the context of writing new research papers.

In recent years, AI's potential to streamline research processes has garnered increasing attention \cite{xu2021artificial}. For instance, expediting literature reviews or illustrating how AI can be used to generate hypotheses \cite{karp1991artificial}. Building upon this growing body of work, our paper examines the capacity of generative AI to serve not just as a tool for operational efficiency, but also as a catalyst for intellectual creativity and innovation in research \cite{lee2023can}.

The integration of Generative Artificial Intelligence (AI) in the realm of academic research, tried by the Galactica model without success \cite{snoswell2022galactica}, particularly in computer science, marks a significant stride towards a more efficient and innovative future. Embracing a techno-optimistic perspective \cite{konigs2022techno}, this paper advocates for the utilization of generative AI as a transformative tool, being an assistant in research paper writing. Generative AI, with its advanced capabilities in data analysis, text generation, and knowledge synthesis, offers an unparalleled opportunity to augment the intellectual creativity and productivity of researchers. By automating routine aspects of writing and data handling, it frees researchers to focus on the more nuanced and creative aspects of their work. This synergy between human intellect and AI's processing power not only enhances the quality and efficiency of research outputs but also propels the frontiers of scientific inquiry, fostering a new era of discovery in computer science. But we also need to be aware about the limitations of current generative AI, that makes it only able to be a research assistant but not a research machine, as for instance its understanding limitations \cite{garrido2022artificial}, its inability to perceive the real world \cite{garrido2023can} and its fundamental biases and misinformation issues \cite{sison2023chatgpt}.  

This paper is organized as follows. We begin with a section called Best Uses of Generative AI for Computer Science Research, where we delve into the core applications of generative AI in the field of computer science research. This section provides a comprehensive analysis of how generative AI tools, particularly ChatGPT, can be utilized to enhance various aspects of computer science research. It covers a range of applications from brainstorming and drafting academic papers to generating synthetic data and aiding in complex text analysis. We continue with a small section where we highlight uses that we do not recommend to apply generative AI. Finally, the paper concludes with a Conclusions and Further Work section that synthesizes the key findings of our research, discussing the implications and potential impact of generative AI in computer science research. We also outline areas for future research, suggesting directions for further exploration and investigation in the field of generative AI.

\section{Best uses of Generative AI for computer science research}
This section delves into the profound impact of generative AI in various facets of computer science research, highlighting its transformative potential. From augmenting the brainstorming process to refining research methodologies, we explore how generative AI not only streamlines the research process but also opens new horizons for innovative inquiry. Each subsection will provide insights into specific use cases, illustrating the diverse and significant contributions of generative AI in advancing the field of computer science research.

\subsection{Brainstorming Research Ideas}

The application of generative AI in brainstorming and formulating research ideas marks a significant advancement in the field of computer science. The initial phase of any research project is critical, as it involves the generation of innovative and feasible ideas. In this phase, generative AI emerges as a crucial tool, enhancing the creative process and expanding the realm of possibilities for researchers.

Generative AI, with its extensive data processing capabilities and access to a vast array of information, can suggest a wide range of potential research topics. These suggestions are grounded in data-driven insights, providing researchers with novel and diverse perspectives that may not have been considered otherwise. This broadens the scope of research possibilities, encouraging out-of-the-box thinking and innovation.

Human researchers, while capable of profound creativity, can be limited by cognitive biases and knowledge constraints. Generative AI, devoid of such biases, can introduce novel ideas and approaches, thereby diversifying the thought process. This helps in overcoming potential blind spots in the ideation phase, leading to more comprehensive and varied research proposals.

Furthermore, generative AI's ability to analyze and synthesize information across various fields can facilitate interdisciplinary research. By identifying and combining concepts from different disciplines, AI can assist researchers in crafting projects that are not only innovative but also have broader implications and applications.

\subsection{Translation and Styling of Academic Papers}

Generative AI is not only an asset for idea generation but also a vital tool in bridging language barriers \cite{klimova2023neural} and enhancing the stylistic quality of academic papers. The necessity for accurate translation in academic research cannot be overstated, especially in an era where collaboration transcends geographical and linguistic boundaries. Generative AI, with its advanced language models \cite{zhao2023survey}, provides highly accurate translation services, enabling researchers to access and contribute to literature in multiple languages. This democratization of knowledge is crucial for the global dissemination and advancement of research.

Beyond translation, generative AI significantly contributes to the styling and formatting of academic papers. Adherence to specific publication guidelines and stylistic norms can be a daunting task for researchers. AI tools, trained on a myriad of academic writing formats, assist in ensuring that manuscripts comply with the required stylistic and formatting standards of various journals, thereby streamlining the submission process and enhancing the likelihood of publication acceptance.

Moreover, AI-driven tools offer suggestions to improve the readability and coherence of academic texts. By analyzing sentence structure, coherence, and overall flow, these tools provide constructive feedback, enabling researchers to refine their manuscripts into more effective and impactful academic papers.

\subsection{State-of-the-art Assistant}
One of the key functions of AI in this context is automating the process of literature review and analysis. By swiftly scanning through vast databases of published work with plugins, AI can identify and summarize key findings, theories, and methodologies relevant to a specific research area. This not only saves considerable time but also ensures that the literature review is exhaustive and up-to-date.

Furthermore, AI is capable of detecting emerging trends and research gaps by analyzing patterns and frequencies of topics in the literature. This insight is invaluable for researchers aiming to position their work within the current research landscape and to contribute novel perspectives or solutions to existing challenges.

AI tools can also assist in broadening the scope of the SOTA section by suggesting references across disciplines that might be relevant. This interdisciplinary approach enriches the research, offering a more holistic view and potentially revealing unexplored connections.

\subsection{Drafting Abstracts and Conclusions}
Generative AI's role in drafting abstracts and conclusions of academic papers is an area where its impact is particularly noteworthy. Abstracts and conclusions are critical components of research papers, requiring a concise yet comprehensive summary of the research and its findings.

For abstract creation, generative AI can process the entire content of a paper to extract key points, ensuring that the abstract accurately reflects the core objectives, methods, results, and implications of the research. This not only aids in maintaining brevity and clarity but also ensures that all vital information is included, which is essential for readers who often rely on the abstract to gauge the relevance of the paper.

In drafting conclusions, AI assists in synthesizing the research findings, drawing connections to the research questions and objectives stated earlier in the paper. It can suggest insightful ways to discuss the implications of the findings, potential applications, and future research directions. This helps in providing a powerful and impactful closure to the paper, which is crucial for leaving a lasting impression on the reader.

Moreover, generative AI tools ensure consistency and coherence between the abstract, the body of the paper, and the conclusions. This is vital in academic writing, as it maintains the integrity and flow of the paper, making it more effective and reader-friendly.

Additionally, AI can be tailored to adhere to the specific requirements of different academic journals, which often have varying guidelines for abstracts and conclusions. This customization saves time for researchers and increases the likelihood of paper acceptance.

\subsection{Code interpreter and data analysis}
Generative AI significantly contributes to computer science research as a code interpreter and data analysis tool \cite{feng2023investigating}. It simplifies complex code interpretation tasks and streamlines data analysis, enhancing research efficiency and accuracy.

AI's ability to interpret and explain code is invaluable, particularly when dealing with large codebases or unfamiliar programming languages. It can provide insights into code functionality, suggest optimizations, and identify potential errors, making the development process more efficient.

In data analysis, AI algorithms can quickly process large datasets, perform statistical analyses, and identify patterns or anomalies. This capability is crucial for researchers dealing with big data, as it allows them to focus on interpretation and application of the findings rather than the intricacies of data processing.

Overall, generative AI as a code interpreter and data analysis tool is indispensable in modern computer science research, offering significant time savings and enhanced accuracy in these technical tasks.

\subsection{Simplification of Complex Texts}
This application of AI is particularly beneficial in interpreting technical documents, academic papers, and data-rich texts. AI tools are adept at breaking down technical jargon and complex concepts into simpler, more understandable language. This is especially useful for researchers who may be delving into interdisciplinary fields or reviewing literature outside their immediate area of expertise. The simplification process also enhances the accessibility of scientific communication, allowing a broader audience to engage with and understand complex research findings. This democratization of knowledge is crucial in a field that thrives on collaborative and cross-disciplinary efforts.

Additionally, generative AI aids in interpreting data-intensive texts, such as research papers with dense statistical information or technical reports. By summarizing and clarifying key points, AI tools help researchers quickly grasp the essence of the text, facilitating a more efficient review process.

\subsection{Suggestion of Academic Journals}
Generative AI significantly aids in the suggestion of appropriate academic journals for research paper submissions, a task crucial for the dissemination of research findings. This application of AI streamlines the publication process and enhances the visibility of research.

AI algorithms analyze the content of a research paper, including its topics, methodologies, and findings, to suggest journals where the paper would be a good fit. This matching process takes into account the scope, audience, and impact factor of potential journals, thereby increasing the likelihood of paper acceptance.

The use of AI for journal suggestion not only saves researchers time in identifying suitable publication venues but also strategically positions their work in the most relevant and impactful journals. This optimization is critical in a competitive academic landscape where publication success is highly valued.

Generative AI tools are also capable of staying updated with the latest trends and changes in academic publishing, including new journals, shifting focus areas, and evolving submission guidelines. This ensures that researchers are always equipped with current and relevant information for their publication strategies.

\subsection{Synthetic data generation}
Generative AI excels in the creation of synthetic data, a capability crucial for research, especially in scenarios where real data may be limited, sensitive, or unavailable. This aspect of AI facilitates the testing of hypotheses and models in a controlled, yet realistic, environment.

By generating realistic datasets, AI enables researchers to bypass constraints related to data privacy and availability, ensuring robust testing without compromising real-world data integrity.

AI's ability to produce synthetic data tailored to specific research requirements ensures versatility across various fields within computer science, aiding in diverse experimental and modeling needs.

\subsection{Methodologist}
Generative AI serves as an effective methodologist in computer science research, offering guidance on research design and methodology selection. This role is pivotal for ensuring the validity and reliability of research findings.

AI tools can suggest appropriate research methodologies based on the research question, objectives, and available data, ensuring that the chosen methods are well-suited to the study's goals.

The involvement of AI in methodological decisions contributes to the rigor and scientific soundness of research projects, leading to more robust and credible outcomes.

\subsection{Research mentor}
Generative AI acts as a research mentor, providing invaluable guidance and support throughout the research process. This role is especially beneficial for novice researchers or those venturing into new areas of computer science.

AI tools can offer step-by-step guidance, from formulating research questions to data collection and analysis. This mentorship ensures that researchers follow a structured and logical approach to their studies.

AI can review ongoing work and provide feedback or suggestions for improvement, similar to the role of a human mentor. This ongoing support enhances the quality and coherence of research endeavors.

\subsection{Article Quality Evaluator}
Generative AI plays a crucial role as an evaluator of article quality, ensuring that research papers meet high standards of academic excellence. This application of AI is vital in the process of refining and validating research work before submission or publication.

AI tools are equipped to assess the structural and content quality of research papers. They analyze the logical flow, coherence, and completeness of the arguments presented. By checking for clarity, relevance, and depth in the content, AI helps in ensuring that the paper effectively communicates its research findings and contributes meaningfully to the field.

Generative AI can pinpoint areas in the manuscript that may require further development or clarification. This includes suggesting enhancements in the presentation of data, refinement of arguments, or improvement in the overall narrative. Such detailed feedback is instrumental in elevating the overall quality of the paper.

Additionally, AI tools can verify whether the paper adheres to specific journal or conference submission guidelines, including formatting, citation styles, and word limits. This compliance check is essential for a smooth submission process and reduces the likelihood of rejection due to non-adherence to guidelines.

Another significant aspect is the AI's ability to conduct plagiarism checks, ensuring the originality and integrity of the research work. This is a critical step in maintaining the ethical standards of academic research.

AI can also perform a comparative analysis of the paper with existing literature to assess its novelty and significance within the field. This analysis helps in positioning the paper in the context of ongoing research and highlights its unique contributions.

\subsection{Summarizing Texts for Length Adaptation}

Generative AI significantly aids in summarizing texts to meet specific length requirements \cite{widyassari2022review}, a critical task in academic writing where conciseness and adherence to guidelines are essential. AI's ability to distill complex information into shorter formats without losing key insights is invaluable for researchers.

AI algorithms can efficiently condense content, ensuring that the summarized version maintains the essence and critical points of the original text. This is particularly useful for abstract writing, executive summaries, or when adapting full-length articles into shorter communication pieces.

This tool is also helpful in adhering to strict word count limits set by journals or conferences, allowing researchers to focus on the quality of content rather than the challenge of meeting length constraints.

\subsection{Critical Analysis and Counterargument Formulation}
Generative AI excels in the role of a critical analyst, offering insightful counterarguments and critiques. This function is crucial in academic research, where rigorous debate and the challenging of ideas are fundamental to scientific advancement.

AI tools can analyze a research paper and identify potential weaknesses or areas for further exploration. This critical analysis aids researchers in fortifying their arguments and anticipating possible counterpoints, leading to more robust and defensible research outcomes.

Moreover, generative AI can generate constructive counterarguments, providing a simulated peer review experience. This helps researchers in preemptively addressing potential criticisms and refining their arguments, thereby enhancing the academic rigor of their work.

By presenting alternative viewpoints and challenging the prevailing assumptions, AI encourages a more comprehensive and multifaceted exploration of research topics. This not only strengthens the research itself but also contributes to a broader understanding of the subject matter.

\section{Not recommended uses of Generative AI}
While Generative Artificial Intelligence (AI) has shown remarkable potential in various aspects of computer science research, it is crucial to recognize scenarios where its application may be inappropriate or counterproductive. This awareness helps maintain the integrity and ethical standards of research, ensuring that reliance on AI is both responsible and judicious.

One of the primary areas where generative AI should not be over-relied upon is in substituting human critical thinking and decision-making. While AI can provide data-driven insights and suggestions, the nuanced understanding, ethical considerations, and complex judgments inherent in research should remain the domain of human researchers. Over-reliance on AI for such decisions may lead to a lack of critical engagement with the research topic and potential oversight of ethical implications.

The use of AI in the processing of sensitive or confidential data without proper safeguards is another area of concern. AI systems, unless specifically designed for secure data handling, may pose risks to data privacy and confidentiality. Researchers must be cautious in employing AI for tasks involving sensitive data, ensuring compliance with ethical standards and legal regulations.

Generative AI should also not be solely relied upon for the creation of original research ideas or content. While AI can assist in brainstorming and drafting, the core ideas and arguments should originate from the researcher to maintain the authenticity and originality of the research. Overdependence on AI for content creation risks producing research that lacks depth, originality, and personal insight.

Relying solely on AI for the quality evaluation of academic papers is another area where caution is advised. AI can provide initial assessments regarding structure, coherence, and adherence to guidelines, but it cannot fully appreciate the subtleties of academic argumentation or the theoretical significance of research. Human oversight remains essential to ensure the high quality and academic rigor of research publications.

In sum, while generative AI offers numerous advantages in computer science research, it is imperative to recognize its limitations and areas where its use is not recommended. Balancing AI's capabilities with human oversight, ethical considerations, and a commitment to originality and critical thinking is key to harnessing its potential responsibly and effectively in the research domain.

\section{Conclusions and Further Work}
In conclusion, this paper has explored the expansive and impactful uses of Generative Artificial Intelligence (AI) in the field of computer science research, highlighting its potential to revolutionize various aspects of academic work. From aiding in brainstorming research ideas, translating and styling academic papers, to assisting in the synthesis of state-of-the-art sections \cite{carrera2022conduct}, AI proves to be an invaluable asset. Its capabilities extend to simplifying complex texts, suggesting suitable academic journals, creating synthetic data, and acting as a methodologist and research mentor. Furthermore, AI's role in evaluating article quality, summarizing texts, and formulating critical analyses and counterarguments, underscores its versatility and efficiency in enhancing academic research.

However, alongside these beneficial uses, the paper also delves into the "not recommended uses of generative AI for computer science research," cautioning against over-reliance on AI for tasks where human intuition, ethical considerations, and complex decision-making are paramount. This balanced approach is crucial for harnessing AI's capabilities responsibly and effectively.

Looking ahead, several areas warrant further exploration. Firstly, the ethical implications and challenges posed by the increasing integration of AI in research need comprehensive examination. This includes issues of data privacy, intellectual property, and the potential for AI to perpetuate biases.

Secondly, the development of more sophisticated AI models that can better understand and mimic human creativity and critical thinking in research contexts presents an exciting avenue for future work. This advancement could further enhance AI's utility in more nuanced aspects of research.

Lastly, exploring the integration of AI in interdisciplinary research, particularly how it can bridge gaps between computer science and other fields, could lead to groundbreaking discoveries and innovations.

\bibliographystyle{acm}

\bibliography{main}

\end{document}